\documentclass{article}
\usepackage[preprint]{neurips_2020}

\usepackage[utf8]{inputenc} 
\usepackage[T1]{fontenc}    
\usepackage{hyperref}       
\usepackage{url}            
\usepackage{booktabs}       
\usepackage{amsfonts}       
\usepackage{nicefrac}       
\usepackage{microtype}      
\usepackage{dsfont}
\usepackage{bm}
\usepackage{mathrsfs}
\usepackage{amsmath,mathtools}
\usepackage{amsfonts}
\usepackage{amsthm}
\usepackage{amssymb}
\usepackage{xcolor}
\usepackage{setspace}
\usepackage{float}
\usepackage{url}
\usepackage{multicol}
\usepackage{subfigure}
\usepackage[]{algorithm2e}

\mathtoolsset{showonlyrefs}

\title{A block coordinate descent optimizer for classification problems exploiting convexity}

\author{
 Ravi G. Patel\\
 Center for Computing Research, Sandia National Laboratories\\
 \texttt{rgpatel@sandia.gov}\\
 \And
 Nathaniel A. Trask\\
 Center for Computing Research, Sandia National Laboratories\\
 \texttt{natrask@sandia.gov}\\
 \And
 Mamikon A. Gulian\\
 Center for Computing Research, Sandia National Laboratories\\
 \texttt{mgulian@sandia.gov}
 \And
 Eric C. Cyr\\
 Center for Computing Research, Sandia National Laboratories\\
 \texttt{eccyr@sandia.gov}\\
}

\begin{document}

\maketitle

\begin{abstract}
Second-order optimizers hold intriguing potential for deep learning, but suffer from increased cost and sensitivity to the non-convexity of the loss surface as compared to gradient-based approaches.
We introduce a coordinate descent method to train deep neural networks for classification tasks that exploits global convexity of the cross-entropy loss in the weights of the linear layer. Our hybrid Newton/Gradient Descent (NGD) method is consistent with the interpretation of hidden layers as providing an adaptive basis and the linear layer as providing an optimal fit of the basis to data. By alternating between a second-order method to find globally optimal parameters for the linear layer and gradient descent to train the hidden layers, we ensure an optimal fit of the adaptive basis to data throughout training. The size of the Hessian in the second-order step scales only with the number weights in the linear layer and not the depth and width of the hidden layers; furthermore, the approach is applicable to arbitrary hidden layer architecture. Previous work applying this adaptive basis perspective to regression problems demonstrated significant improvements in accuracy at reduced training cost, and this work can be viewed as an extension of this approach to classification problems. We first prove that the resulting Hessian matrix is symmetric semi-definite, and that the Newton step realizes a global minimizer. By studying classification of manufactured two-dimensional point cloud data, we demonstrate both an improvement in validation error and a striking qualitative difference in the basis functions encoded in the hidden layer when trained using NGD. Application to image classification benchmarks for both dense and convolutional architectures reveals improved training accuracy, suggesting possible gains of second-order methods over gradient descent. A Tensorflow implementation of the algorithm is available at \url{github.com/rgp62/}.
  
\end{abstract}

\section{A Newton/gradient coordinate descent  optimizer for classification}
Denote by $\mathcal{D} = \left\{(\mathbf{x}_i,\mathbf{y}_i)\right\}_{i=1}^{N_\text{data}}$ data/label pairs, and consider the following class of deep learning architectures: 
\begin{equation}\label{eq:architecture}
    \mathcal{L}(\mathbf{W},\xi,\mathcal{D}) = \sum_{(\mathbf{x}_i,\mathbf{y}_i) \in \mathcal{D}} \mathcal{L}_{\text{CE}}(\cdot;\mathbf{y}_i) \circ \mathcal{F}_{\text{SM}} \circ \mathcal{F}_{\text{LL}} (\cdot; \mathbf{W}) \circ \mathcal{F}_{\text{HL}} (\mathbf{x}_i; \xi),
\end{equation}
where $\mathcal{L}_{\text{CE}},\mathcal{F}_{\text{SM}},\mathcal{F}_{\text{LL}}$ and $\mathcal{F}_{\text{HL}}$ denote a cross-entropy loss, softmax layer, linear layer, and hidden layer, respectively. We denote linear layer weights by $\mathbf{W}$, and consider a general class of hidden layers (e.g. dense networks, convolutional networks, etc.), denoting associated weights and biases by the parameter $\xi$. The final three layers are expressed as
\begin{equation}\label{eq:final_layers}
    \mathcal{L}_{\text{CE}}(\mathbf{x};\mathbf{y}) = -{\sum_{i=1}^{N_c} {y}^i \log {x}^i}; \quad
    \mathcal{F}^i_{\text{SM}}(\mathbf{x}) = \frac{\exp (-x_i)}{\sum_{j=1}^{N_c}\exp(-x_j)};\quad
    \mathcal{F}^i_{\text{LL}}(\mathbf{x}) = {\mathbf{W}} \mathbf{x}
\end{equation}
and map $\mathcal{F}_{\text{HL}}:\mathbb{R}^{N_{\text{in}}}\rightarrow \mathbb{R}^{N_{\text{basis}}}$;  $\mathcal{F}_{\text{LL}}:\mathbb{R}^{N_{\text{basis}}}\rightarrow \mathbb{R}^{N_{\text{classes}}}$; $\mathcal{F}_{\text{SM}}:\mathbb{R}^{N_{\text{classes}}}\rightarrow \mathbb{R}^{N_{\text{classes}}}$; and $\mathcal{L}_{\text{CE}}:\mathbb{R}^{N_{\text{classes}}}\rightarrow \mathbb{R}$. Here, $N_{\text{basis}}$ is the dimension of the output of the hidden layer; this notation is explained in the next paragraph. The standard classification problem is to solve
\begin{equation}\label{eq:fullLoss}
    \left(\mathbf{W}^*,\xi^*\right) = \underset{\mathbf{W},\xi}{\text{argmin}}\, \mathcal{L}(\mathbf{W},\xi,\mathcal{D}).
\end{equation}
The recent work by~\citet{cyr2019robust} performed a similar partition of weights into linear layer weights $\mathbf{W}$ and hidden layer weights $\xi$ for regression problems. Two important observations were made using
this decomposition. First, the output of the hidden layers can be treated as an adaptive basis with the learned weights $\mathbf{W}$ corresponding to the coefficients producing the prediction. Second, holding $\xi$ fixed leads to a linear least squares problem for the basis coefficients $\mathbf{W}$ that can be solved for a global minimum. This work builds on these two observations for classification problems. 
The output of the hidden layers $\mathcal{F}_{\text{HL}}$ defines a basis 
\begin{equation} 
\Phi_\alpha(\cdot,\xi): \mathbb{R}^{N_{\text{in}}} \rightarrow \mathbb{R} \mbox{ for } \alpha = 1 \ldots N_{\text{basis}}
\label{eq:adapt-basis}
\end{equation}
where $\Phi_\alpha(x,\xi)$ is row $\alpha$ of $\mathcal{F}_{\text{HL}}(x,\xi)$.
Thus the input to the softmax classification layer are $N_{\text{classes}}$ functions, each defined using the adaptive basis $\Phi_\alpha$ and a single row of the weight matrix $\mathbf{W}$.
The crux of this approach to classification is the observation that 
for all $\xi$, the function
\begin{equation}\label{eq:S_def}
    \mathcal{S}(\mathbf{W}, \mathcal{D}) = \mathcal{L}(\mathbf{W},\xi, \mathcal{D})
\end{equation}
is convex with respect to $\mathbf{W}$,
and therefore the global minimizer
\begin{equation}\label{eq:WLoss}
    \mathbf{W}^* = \underset{\mathbf{W}}{\text{argmin}}\, \mathcal{S}(\mathbf{W}, \mathcal{D})
\end{equation}
may be obtained via Newton iteration with line search. 
In  Sec. \ref{sec:algo}, we introduce a coordinate-descent optimizer that alternates between a globally optimal solution of \eqref{eq:WLoss} and a gradient descent step minimizing \eqref{eq:fullLoss}. Combining this with the interpretation of the hidden layer as providing a data-driven adaptive basis, this ensures that during training the parameters evolve along a manifold providing optimal fit of the adaptive basis to data \citep{cyr2019robust}. We summarize this perspective and relation to previous work in Sec. \ref{sec:literature}, and in Sec. \ref{sec:results} we investigate how this approach differs from stochastic gradient descent (GD), both in accuracy and in the qualitative properties of the hidden layer basis. 

\section{Convexity analysis and invertibility of the Hessian} \label{sec:convexity}
In what follows, we use basic properties of convex functions \citep{boyd2004convex} and the Cauchy-Schwartz inequality \citep{folland1999real} to prove that $\mathcal{S}$ in \eqref{eq:S_def} is convex. Recall that convexity is preserved under affine transformations. We first note that $\mathcal{L}_{\text{LL}}(\mathbf{W};\mathcal{D},\xi)$ is linear.
By \eqref{eq:architecture}, it remains only to show that $\mathcal{L}_{\text{CE}} \circ \mathcal{F}_{\text{SM}}$ is convex. We write, for any data vector $\mathbf{y}$,
\begin{align}
    \mathcal{L}_{\text{CE}} \circ \mathcal{F}_{\text{SM}} (\mathbf{x}; \mathbf{y}) &= - \sum_{i=1}^{N_{\text{classes}}} y_i
    \log\left(\frac{\exp(-x_i)}{\sum_{j=1}^{N_\text{classes}} \exp(- x_j)}\right)\\
    &= \sum_{i=1}^{N_{\text{classes}}} y_i
    x_i - N_{\text{classes}} \log \left( \sum_{i=1}^{N_{\text{classes}}} \exp \left(-x_i \right)\right).
\end{align}
The first term above is affine and thus convex. We prove the convexity of the second term 
${f(\mathbf{x}) := -\log \left( \sum_{i=1}^{N_{\text{classes}}} \exp \left(-x_i \right)\right)}$ by writing
\begin{equation}
    f(\theta \mathbf{x} + (1-\theta)\mathbf{y}) = \log \left( \sum_{i=1}^{N_{\text{classes}}} 
    \left( \exp(-x_i) \right)^\theta 
    \left( \exp(-y_i) \right)^{1-\theta}  
    \right).
\end{equation}
Applying Cauchy-Schwartz with $1/p = \theta$ and $1/q = 1 - \theta$, noting that $1/p + 1/q = 1$, we obtain
\begin{align}
    f(\theta \mathbf{x} + (1-\theta)\mathbf{y}) &\leq 
    \log \left(  
    \left( \sum_{i=1}^{N_{\text{classes}}}\exp(-x_i) \right)^\theta 
    \left( \sum_{i=1}^{N_{\text{classes}}}\exp(-y_i) \right)^{1-\theta}  
    \right)\\
    &= \theta f(\mathbf{x}) + (1-\theta)f(\mathbf{y}).
\end{align}
Thus $f$, and therefore $\mathcal{L}_{\text{CE}} \circ \mathcal{F}_{\text{SM}}$ and $\mathcal{S}$, are convex. As a consequence, the Hessian $H$ of $\mathcal{S}$ with respect to $\mathbf{W}$ is a symmetric positive semi-definite function, allowing application of a convex optimizer in the following section to realize a global minimum.

\section{Algorithm} \label{sec:algo}

Application of the traditional Newton method to the problem \eqref{eq:fullLoss} would require solution of a dense matrix problem of size equal to the total number of parameters in the network. In contrast, we alternate between applying Newton's method to solve only for $\mathbf{W}$ in \eqref{eq:WLoss} and a single step of a gradient-based optimizer for the remaining parameters $\xi$; the Newton step therefore scales with the number of weights ($N_{\text{basis}}\times N_{\text{classes}}$) in the linear layer. Since $\mathcal{S}$ is convex,
Newton's method with appropriate backtracking or trust region may be expected to achieve a global minimizer. We pursue a simple backtracking approach, taking the step direction and size from standard Newton and repeatedly reducing the step direction until the Armijo condition is satisfied, ensuring an adequate reduction of the loss \citep{armijo1966minimization,dennis1996numerical}. For the gradient descent step we apply Adam \citep{kingma2014adam}, although one may apply any gradient-based optimizer; we denote such an update to the hidden layers for fixed $\mathbf{W}$ by the function $\text{GD}(\xi,\mathcal{B},\mathbf{W})$. To handle large data sets, stochastic gradient descent (GD) updates parameters using gradients computed over disjoint subsets $\mathcal{B} \subset \mathcal{D}$ \citep{bottou2010large}. 
To expose the same parallelism, we apply our coordinate descent update over the same batches by solving \eqref{eq:WLoss} restricted to $\mathcal{B}$. Note that this implies an optimal choice of $\mathbf{W}$ over $\mathcal{B}$ only. We summarize the approach in Alg. \ref{thealg}. 

\RestyleAlgo{plainruled}
\begin{algorithm}[t]
 \KwData{batch $\mathcal{B}\subset \mathcal{D},\xi_{\text{old}},\mathbf{W}_{\text{old}},\alpha,\rho$}
 \KwResult{$\xi_{\text{new}}, \mathbf{W}_{\text{new}}$}
 \For{$j \in \{1,...,\texttt{\textup{newton\_steps}}\}$}{
      Compute gradient $G = \nabla_{\mathbf{W}} S(\mathbf{W}_{\text{old}},\mathcal{B}) $ and Hessian $H = \nabla_{\mathbf{W}} \nabla_{\mathbf{W}} S(\mathbf{W}_{\text{old}},\mathcal{B})$ \;
      Solve $H\bm{s} = -G$\;
      $\mathbf{W}^{\dagger} \gets \mathbf{W}_{\text{old}} + \bm{s}$\;
      $\lambda \gets 1$\;
      \While{$S(\mathbf{W}^{\dagger},\mathcal{B}) > S(\mathbf{W}_{\textup{old}},\mathcal{B}) + \alpha \lambda G \cdot \bm{s}$}{
          $\lambda \gets \lambda \rho$;\\
          $\mathbf{W}^{\dagger} \gets  \mathbf{W}_{\text{old}} + \lambda \bm{s}$\;
      }
  }
  $\mathbf{W}_{\text{new}} \gets \mathbf{W}^\dagger$\;
    ${\xi}_{\text{new}} \gets \mathrm{GD}(\xi_{\text{old}},\mathcal{B},\mathbf{W}_{\text{new}})$\;
 \caption{Application of coordinate descent algorithm for classification to a single batch $\mathcal{B}\subset \mathcal{D}$. For the purposes of this work, we use $\rho = 0.5$ and $\alpha = 10^{-4}$.}
 \label{thealg}
\end{algorithm}
While $H$ and $G$ may be computed analytically from \eqref{eq:final_layers}, we used automatic differentiation for ease of implementation.
The system $H\bm{s} = -G$ can be solved using either a dense or an iterative method. Having proven convexity of $\mathcal{S}$ in \eqref{eq:WLoss}, and thus positive semi-definiteness of the Hessian, we may apply a conjugate gradient method. We observed that solving to a relatively tight residual resulted in overfitting during training, while running a fixed number $N_{cg}$ of iterations improved validation accuracy. Thus, we treat $N_{cg}$ as a hyperparameter in our studies below. We also experimented with dense solvers; due to rank deficiency we considered a pseudo-inverse of the form $H^\dagger = (H + \epsilon I)^{-1}$, where taking a finite $\epsilon>0$ provided similar accuracy gains. We speculate that these approaches may be implicitly regularizing the training. For brevity we only present results using the iterative approach; the resulting accuracy was comparable to the dense solver. In the following section we typically use only a handful of Newton and CG iterations, so the additional cost is relatively small.

We later provide convergence studies comparing our technique to GD using the Adam optimizer and identical batching. We note that a lack of optimized software prevents a straightforward comparison of the performance of our approach vs. standard GD;  while optimized GPU implementations are already available for GD, it is an open question how to most efficiently parallelize the current approach. For this reason we compare performance in terms of iterations, deferring wall-clock benchmarking to a future work when a fair comparison is possible.

\section{Relation to previous works} \label{sec:literature}

We seek an extension of \citet{cyr2019robust}. This work used an adaptive basis perspective to motivate a block coordinate descent approach utilizing a linear least squares solver.
The training strategy they develop can be found under the name of variable projection, and was used to train small networks~\citep{mcloone1998hybrid,pereyra2006variable}.
In addition to the work in \citet{cyr2019robust}, the perspective of neural networks producing an adaptive basis has been considered by several approximation theorists to study the accuracy of deep networks \citep{yarotsky2017error,opschoor2019deep, daubechies2019nonlinear}.  The combination of the adaptive basis perspective combined with the block coordinate descent optimization demonstrated dramatic increases in accuracy and performance in \citet{cyr2019robust}, but was limited to an $\ell_2$ loss. None of the previous works have considered the generalization of this approach to training deep neural networks with a cross-entropy loss typically used in classification as we develop here.

\citet{bottou2018optimization} provides a mathematical summary on the breadth of work on numerical optimizers used in machine learning.
Several recent works have sought different means to incorporate second-order optimizers to accelerate training and avoid issues with selecting hyperparameters and training schedules \citep{osawa2019large,osawa2020scalable,botev2017practical,martens2010deep}. Some pursue a quasi-Newton approach, defining approximate Hessians, or apply factorization to reduce the effective bandwidth of the Hessian \citep{botev2017practical,xu2019newton}. Our work pursues a (block) coordinate descent strategy, partitioning degrees of freedom into sub-problems amenable to more sophisticated optimization \citep{nesterov2012efficiency,wright2015coordinate,blondel2013block}. Many works have successfully employed such schemes in ML contexts (e.g. \citep{blondel2013block,fu1998penalized,shevade2003simple,clarkson2012sublinear}), but they typically rely on stochastic partitioning of variables rather than the partition of the weights of deep neural networks into hidden layer variables and their complement pursued here. The strategy of extracting convex approximations to nonlinear loss functions is classical \citep{bubeck2014convex}, and some works have attempted to minimize general loss functions by minimizing surrogate $\ell_2$ problems \citep{barratt2020least}.

\section{Results} \label{sec:results}
We study the performance and properties of the NGD algorithm as compared to the standard stochastic gradient descent (GD) on several benchmark problems with various architectures. We start by applying dense network architectures to classification in the peaks problem. This allows us to plot and compare the qualitative properties of the basis functions $\Phi_\alpha(\cdot,\xi)$
encoded in the hidden layer \eqref{eq:adapt-basis} when trained with the two methods. We then compare the performance of NGD and GD for the standard image classification benchmarks CIFAR-10, MNIST, and Fashion MNIST using both dense and convolutional (ConvNet) architectures. Throughout this section, we compare performance in terms of iterations of Alg. \ref{thealg} for NGD and iterations of stochastic gradient descent, each of which achieves a single update of the parameters $(\mathbf{W}, \xi)$ in the respective algorithm based on a batch $\mathcal{B}$; this is the number of epochs multiplied by the number of batches.  

\subsection{Peaks problem} \label{sec:peaks}
\begin{figure}
    \centering
    \includegraphics[width=0.8\textwidth]{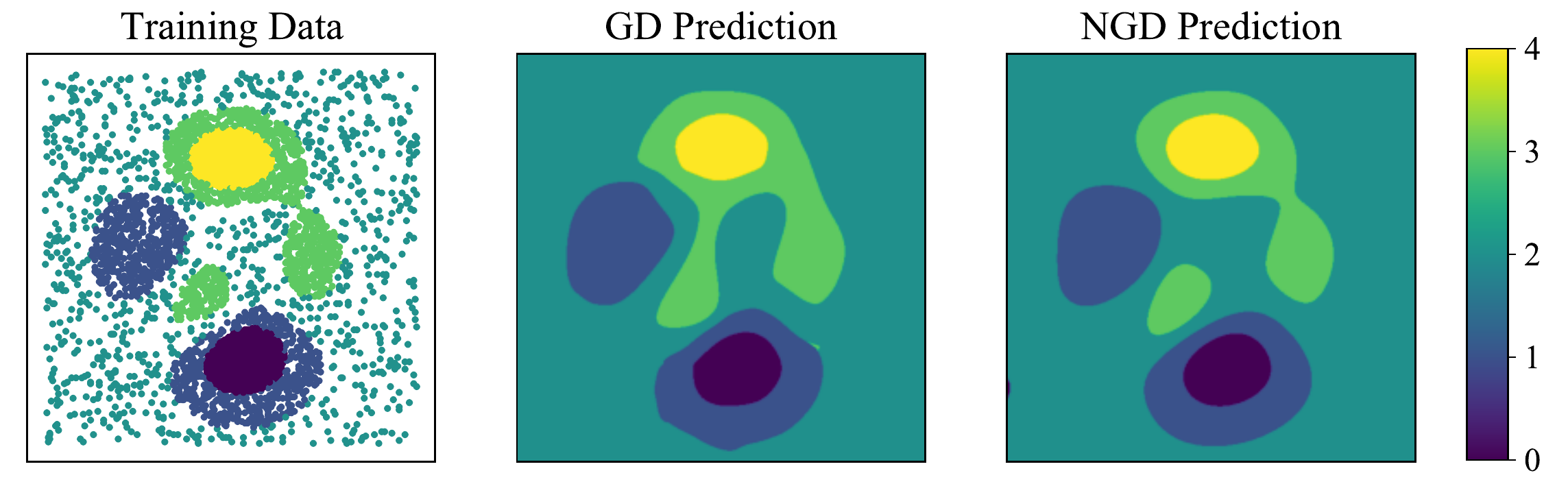} 
    \includegraphics[width=0.4\textwidth]{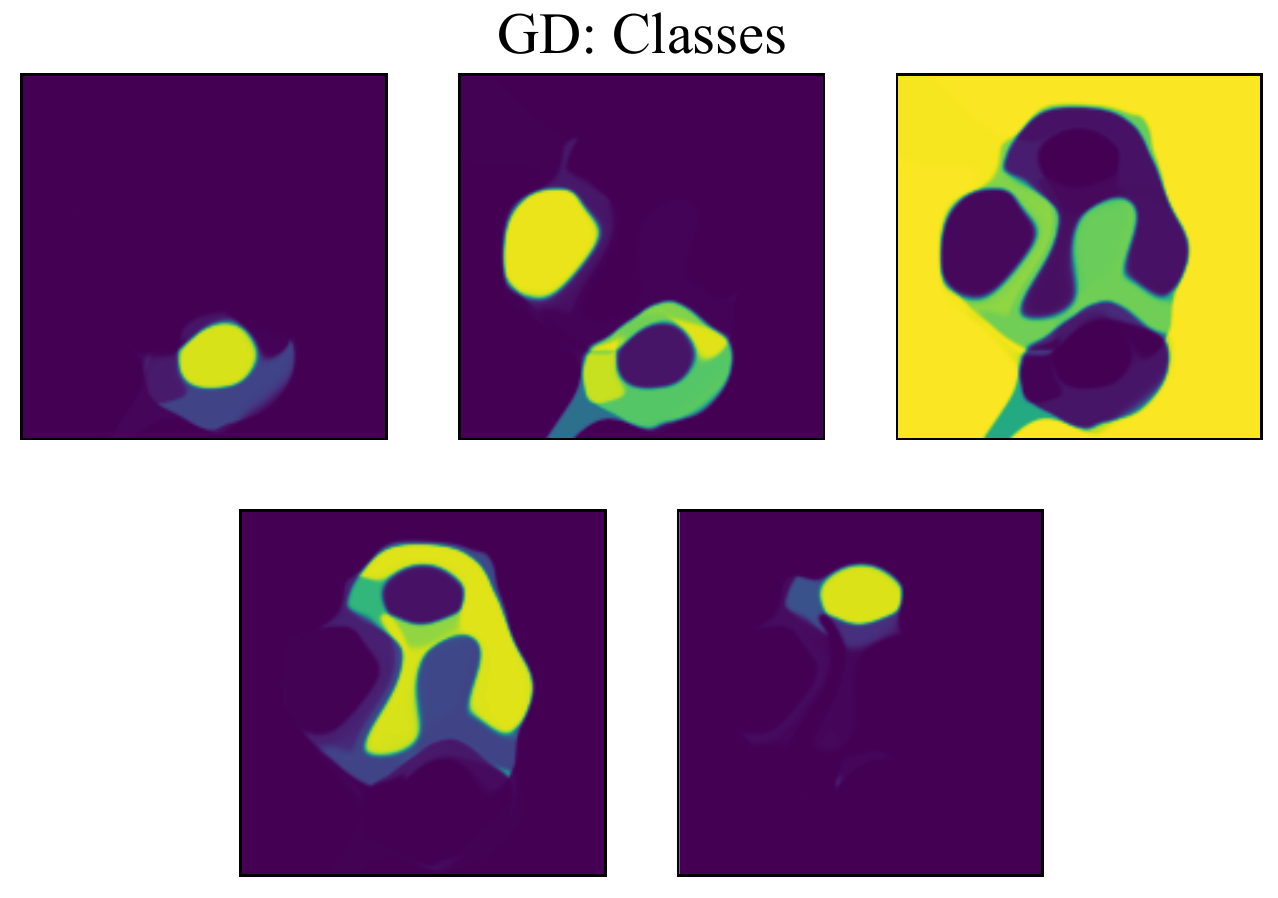}
    \hspace{0.05\textwidth}
    \includegraphics[width=0.4\textwidth]{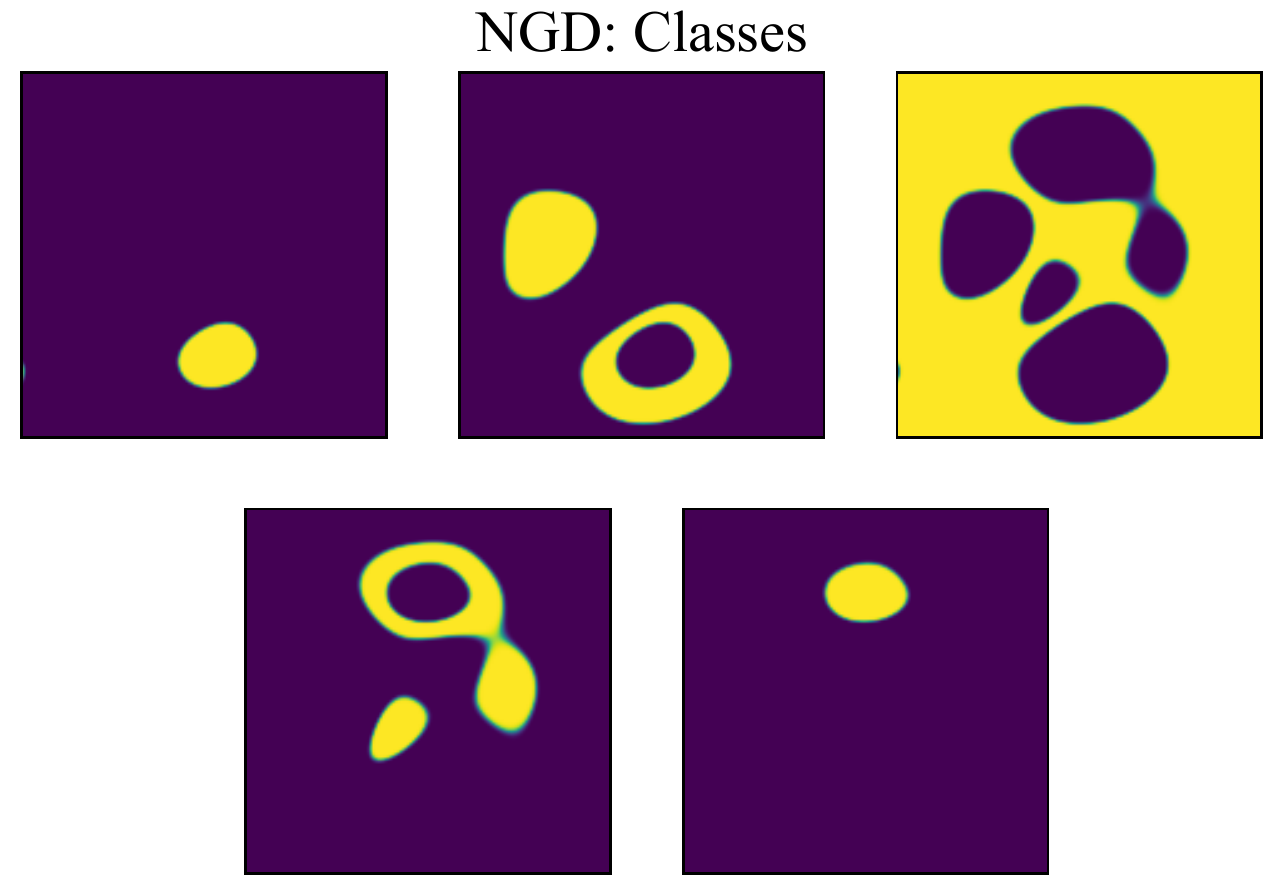}
    \includegraphics[width=0.4\textwidth]{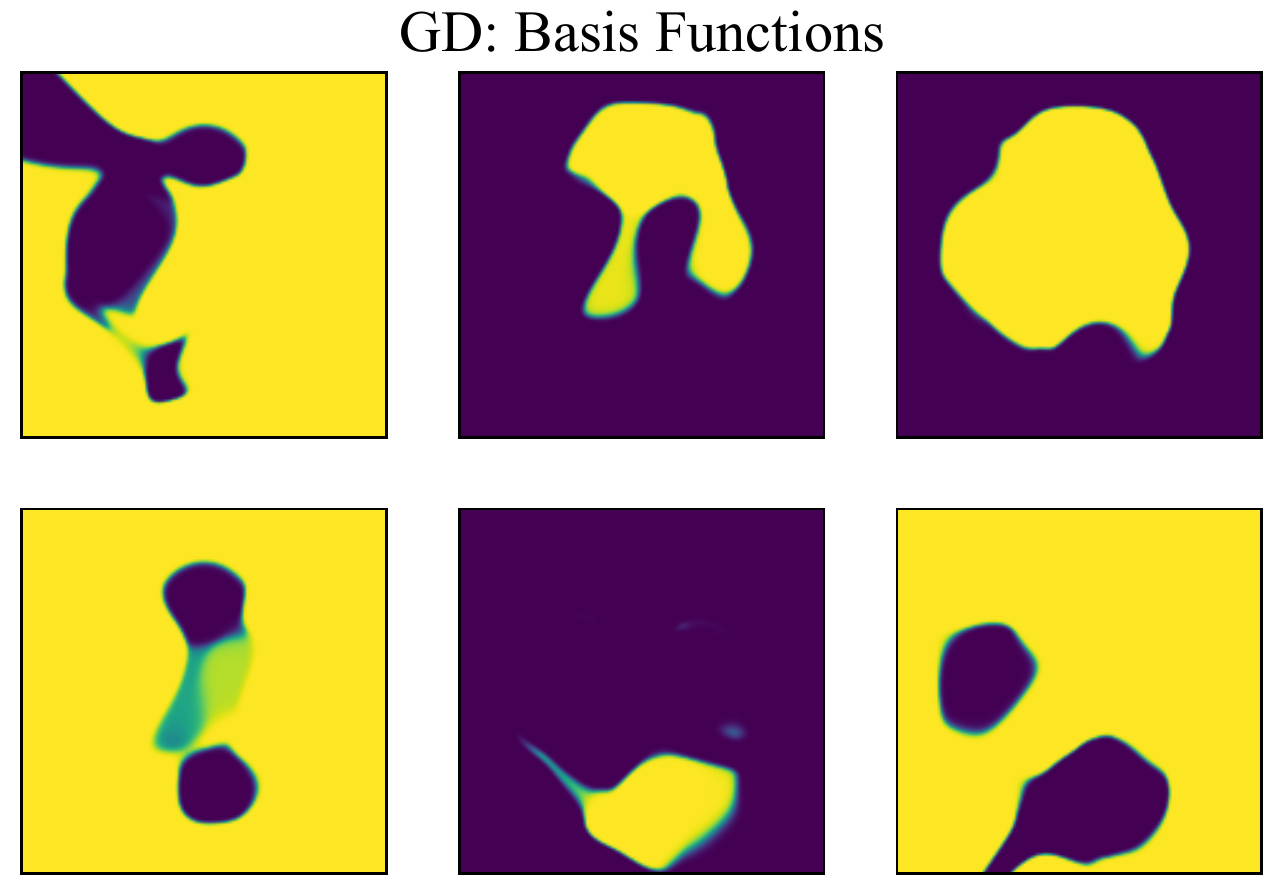}
    \hspace{0.05\textwidth}
    \includegraphics[width=0.4\textwidth]{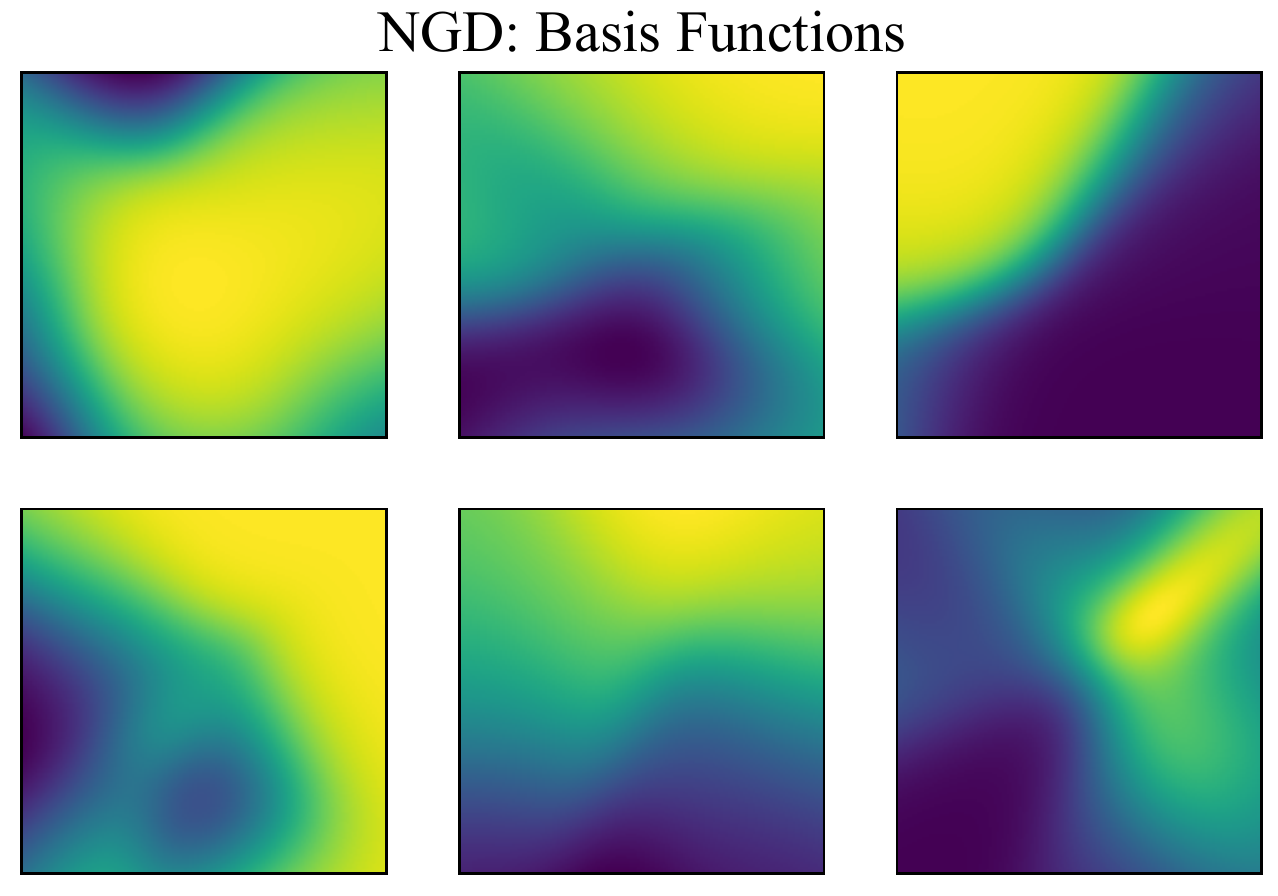}
    \caption{Results for peaks benchmarks, with comparison between NGD and GD on an identical architecture. In this example, GD obtained a training accuracy of 99.3\% and validation accuracy of 96.2\%, while NGD obtained a training accuracy of 99.6\% and validation accuracy of 98.0\%. \textbf{Top:} Training data (\textit{left}), classification by GD (\textit{center}), and classification by NGD (\textit{right}). GD misclassifies large portions of the input space.
    \textbf{Middle:} The linear and softmax layers combine basis functions to assign classification probabilities to each class. The sharp basis learned in GD leads to artifacts and attribution of probability far from the sets (\textit{left}) while diffuse basis in NGD provides a sharp characterization of class boundaries (\textit{right}).
    \textbf{Bottom:} Adaptive basis encoded by hidden layer, as learnt by GD (\textit{left}) and NGD (\textit{right}). For GD the basis is sharp, and individual basis functions conform to classification boundaries, while NGD learns a more regular basis. }
    \label{fig:peaks2}
\end{figure}

The peaks benchmark is a synthetic dataset for understanding the qualitative performance of classification algorithms \citep{haber2017stable}. Here, a scattered point cloud in the two- dimensional unit square $[0,1]^2$ is partitioned into disjoint sets. The classification problem is to determine which of those sets a given 2D point belongs to.
The two-dimensional nature allows visualization of how NGD and GD classify data. In particular, plots of both how the nonlinear basis encoded by the hidden layer maps onto classification space and how the linear layer combines the basis functions to assign a probability map over the input space are readily obtained. We train a depth 4 dense network of the form \eqref{eq:architecture} with $N_{\text{in}} = 2$, three hidden layers of width 12 contracting to a final hidden layer of width $N_{\text{basis}} = 6$, with $\tanh$ activation and batch normalization, and $N_{\text{classes}} = 5$ classes.
As specified by \citet{haber2017stable}, $5000$ training points are sampled from $[0,1]^2$. The upper-left most image in Figure~\ref{fig:peaks2} shows the sampled data points with their observed classes. For the peaks benchmark we use a single batch containing all training points, i.e. $\mathcal{B} = \mathcal{D}$. The NGD algorithm uses $5$ Newton iterations per training step with $3$ CG iterations approximating the linear solve. The learning rate for Adam for both NGD and GD is $10^{-4}$.

\begin{figure}
    \centering
    \includegraphics[width=0.49\textwidth]{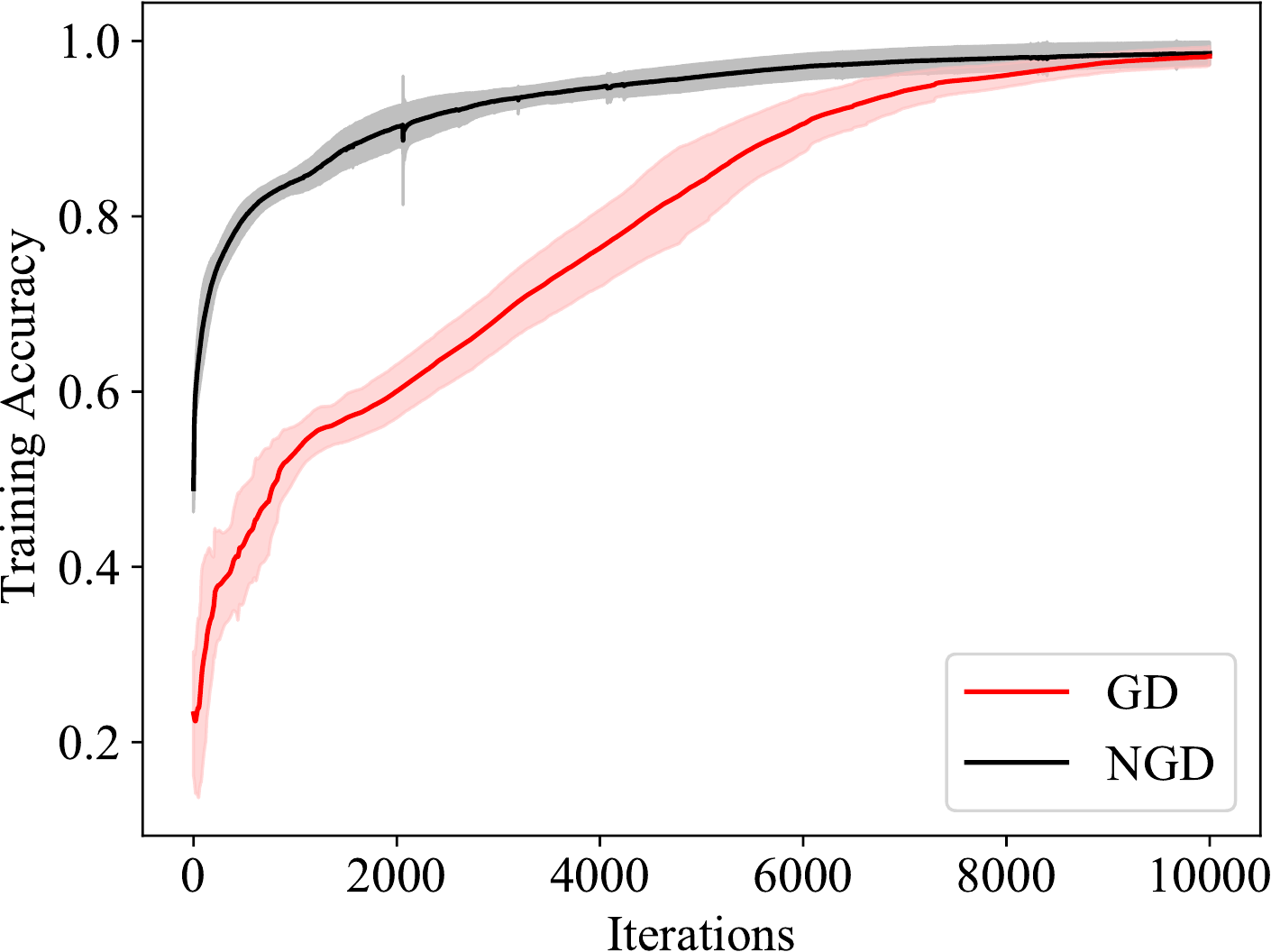}
    \includegraphics[width=0.49\textwidth]{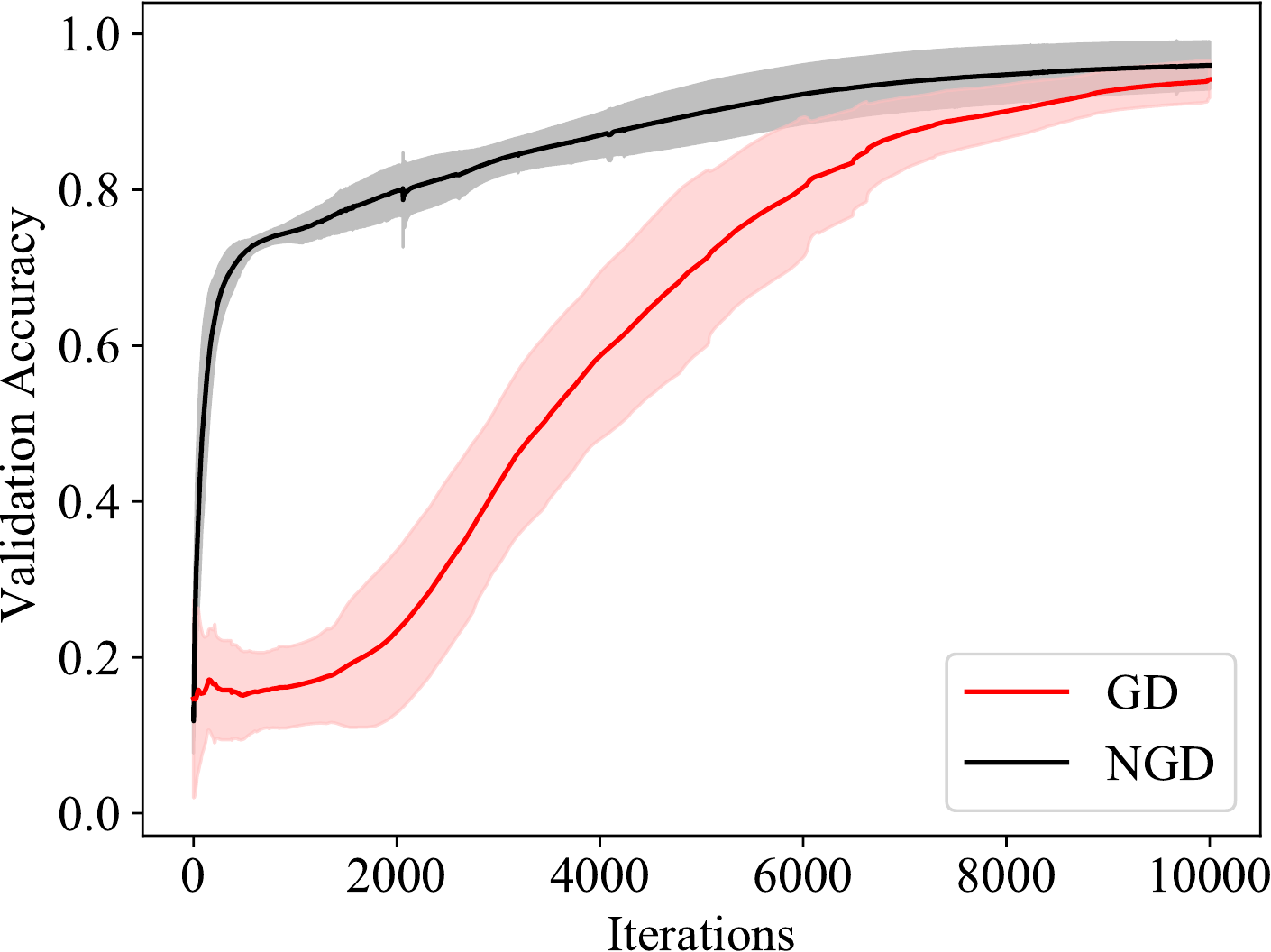}
    \caption{The training (\textit{left}) and validation (\textit{right}) accuracy for the peaks problem for both gradient descent (GD) and the Newton/gradient descent (NGD) algorithm. The solid lines represent the mean of 16 independent runs, and the shaded areas represent the mean $\pm$ one standard deviation.}
    \label{fig:peaks1}
\end{figure}

Figure \ref{fig:peaks1} demonstrates that for an identical architecture, NGD provides a rapid increase in both training and validation accuracy over GD after a few iterations. For a large number of iterations both approaches achieve comparable training accuracy, although NGD generalizes better to the validation set. 
The improvement in validation accuracy is borne out in Figure \ref{fig:peaks2}, which compares representative instances of training using GD and NGD. 
While a single instance is shown, the character of these results is consistent with other neural networks trained for the Peaks problem in the same way.
The top row illustrates the predicted classes $\text{argmax } [\mathcal{F}_{\text{SM}}(\mathbf{x})] \in \{0,1,2,3,4\}$ for $\mathbf{x} \in [0,1]^2$ and the training data, demonstrating that
the NGD-trained network predicts the class $i=2$ of lowest training point density more accurately than the GD-trained network. 
The remaining sets of images visualize both the classification probability map $[\mathcal{F}_{\text{SM}}(\mathbf{x})]_i$ for $i \in \{0,1,2,3,4\}$ (middle row) and the six basis functions $\Phi_\alpha(\mathbf{x}, \xi)$ (bottom row) learned by each optimizer. 
The difference in the learned bases is striking. 
GD learns a basis that is nearly discontinuous, in which the support of each basis function appears fit to the class boundaries. On the other hand, NGD learns a far smoother basis that can be combined to give sharper class boundary predictions.
This manifests in the resulting probability map assigned to each class; linear combinations of the rougher GD basis results in imprinting and assignment of probability far from the relevant class.
This serves as an explanation of the improved validation accuracy of NGD as compared to GD despite similar final training accuracy. 
The NGD algorithm separates refinement of the basis from the determination of the coefficients. This provides an effective regulation of the final basis, leading to improved generalization.

\subsection{Image recognition benchmarks}

We consider in this section a collection of image classification benchmarks: MNIST \citep{deng2012mnist,grother1995nist}, fashion MNIST \citep{xiao2017fashion}, and CIFAR-10 \citep{krizhevsky2009learning}. We focus primarily on CIFAR-10 due to its increased difficulty;
it is well-known that one may obtain near-perfect accuracy in the MNIST benchmark without sophisticated choice of architecture.
For all problems, we consider a simple dense network architecture to highlight the role of the optimizer, and for CIFAR-10 we also utilize convolutional architectures (ConvNet). This highlights that our approach applies to general hidden layer architectures. Our objective is to demonstrate improvements in accuracy due to optimization with all other aspects held equal. For CIFAR-10, for example, the state-of-the-art requires application of techniques such as data-augmentation and complicated architectures to realize good accuracy; for simplicity we do not consider such complications to allow a simple comparison. The code for this study is provided at \url{github.com/rgp62/}. 

\begin{figure}
    \centering
    \includegraphics[width=1\textwidth]{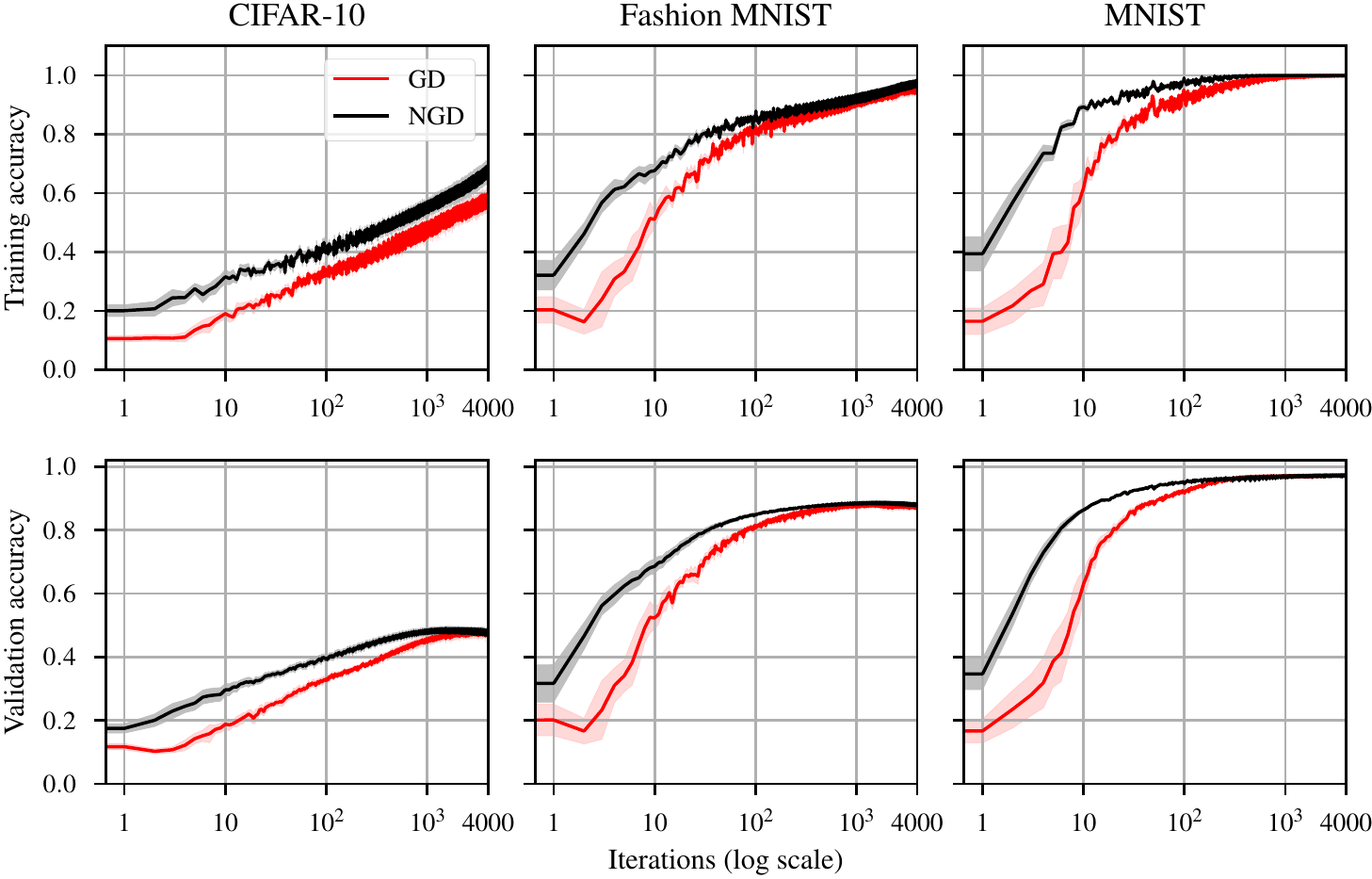}
    \caption{Training accuracy (\textit{top row}) and validation accuracy (\textit{bottom row}) for CIFAR-10, Fashion MNIST, and MNIST datasets. Mean and standard deviation over 10 training runs are shown.}
    \label{fig:three_sets}
\end{figure}

\begin{figure}
    \centering
    \includegraphics[width=5in]{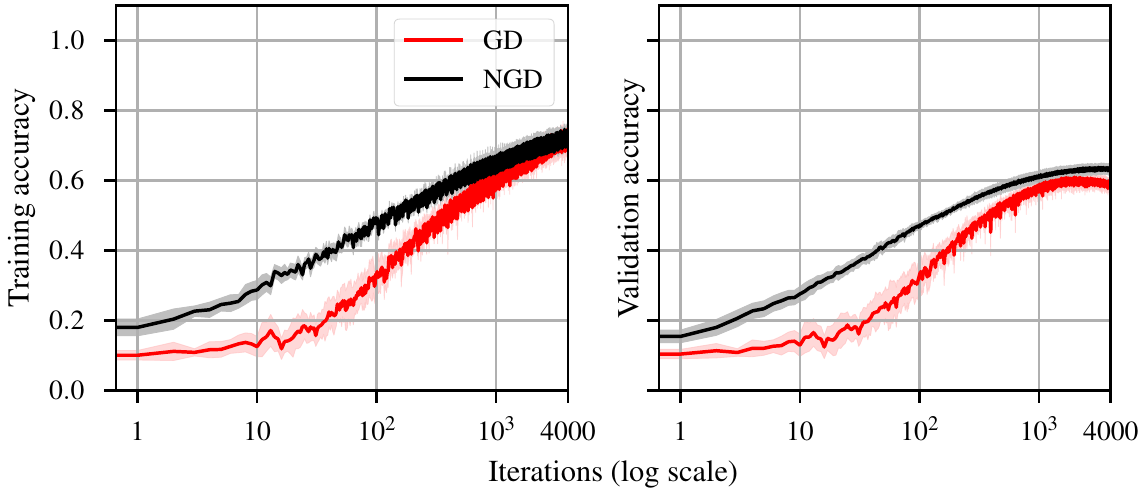}
    \caption{Training accuracy (\textit{left}) and validation accuracy (\textit{right}) for ConvNet architectures. Mean and standard deviation over 10 training runs are shown.}
    \label{fig:convnet}
\end{figure}

For all results reported in this section, we first optimize the hyperparameters listed in Table~\ref{tab:parameters} by maximizing the validation accuracy over the training run. We perform this search using the Gaussian process optimization tool in the scikit-optimize package with default options \citep{scikitopt}. This process is performed for both GD and NGD to allow a fair comparison. The ranges for the search are shown in Table~\ref{tab:parameters} with the optimal hyperparameters for each dataset examined in this study. For all problems we partition data into training, validation and test sets to ensure hyperparameter optimization is not overfitting. For MNIST and fashion MNIST we consider a $50K/10K/10K$ partition, while for CIFAR-10 we consider a $40K/10K/10K$ partition. All training is performed with a batch size of $1000$ over $100$ epochs. For all results the test accuracy falls within the first standard deviation error bars included in Figures \ref{fig:three_sets} and \ref{fig:convnet}.

Figure~\ref{fig:three_sets} shows the training and validation accuracies using the optimal hyperparameters for a dense architecture with two hidden layers of width 128 and 10 and ReLU activation functions. We find for all datasets, NGD more quickly reaches a maximum validation accuracy compared to GD, while both optimizers achieve a similar final validation accuracy. For the more difficult CIFAR-10 benchmark, NGD attains the maximum validation accuracy of GD in roughly one-quarter of the number of iterations. 
In Figure~\ref{fig:convnet}, we use the CIFAR-10 dataset to compare the dense architecture to the following ConvNet architecture,
\begin{equation*}
\begin{aligned}
    &\underset{\textrm{8 channels, 3x3 kernel}}{\textrm{Convolution}} \rightarrow \underset{\textrm{2x2 window}}{\textrm{Max Pooling}} \rightarrow \underset{\textrm{16 channels, 3x3 kernel}}{\textrm{Convolution}} \\
     & \rightarrow  \underset{\textrm{2x2 window}}{\textrm{Max Pooling}} \rightarrow \underset{\textrm{16 channels, 3x3 kernel}}{\textrm{Convolution}} \rightarrow \underset{\textrm{width 64}}{\textrm{Dense}} \rightarrow \underset{\textrm{width 10}}{\textrm{Dense}}
    \end{aligned}
\end{equation*}
where the convolution and dense layers use the ReLU activation function. Again, NGD attains the maximum validation accuracy of GD in one-quarter the number of iterations, and also leads to an improvement of 1.76\% in final test accuracy. This illustrates that NGD accelerates training and can improve accuracy for a variety of architectures. 

\begin{table}
\begin{center}
\begin{tabular}{ c|c|c|c|c|c } 
 Hyperparameter & range & MNIST & Fashion & CIFAR-10 & CIFAR-10 \\
                &       &       & MNIST   &          & ConvNet  \\
 \hline
  & & & & \\
 Learning rate & $[10^{-8},10^{-2}]$ & $10^{-2.81}$ & $10^{-3.33}$ & $10^{-3.57}$ & $10^{-2.66}$ \\
               &                     & $10^{-2.26}$ & $10^{-2.30}$ & $10^{-2.50}$ & $10^{-2.30}$ \\
 \hline
 Adam decay    & $[0.5,1]$  & $0.537$   & $0.756$ & $0.629$  & $0.755$ \\ 
 parameter 1   &            & $0.630$   & $0.657$ & $0.891$  & $0.657$ \\
 \hline
 Adam decay    & $[0.5,1]$  & $0.830$   & $0.808$ & $0.782$  & $0.858$ \\ 
 parameter 2   &            & $0.616$   & $0.976$ & $0.808$  & $0.976$ \\
 \hline
 CG iterations          & $[1,10]$   & $3$   & $1$  & $2$    & $2$ \\
 \hline
 Newton iterations      & $[1,10]$   & $6$   & $5$  & $4$    & $7$ \\
\end{tabular}
\end{center}
\caption{Hyperparameters varied in study (\textit{first column}), ranges considered (\textit{second column}), and optimal values found for MNIST (\textit{third column}), Fashion MNIST (\textit{fourth column}), CIFAR-10 (\textit{fifth column}), and  CIFAR-10 with the ConvNet architecture (\textit{last column}). For the learning rate and the Adam decay parameters, the optimal values for NGD followed by GD are shown. The optimal CG and Newton iterations are only applicable to NGD.}
\label{tab:parameters}
\end{table}

\section{Conclusions} 
The NGD method, motivated by the adaptive basis interpretation of deep neural networks, is a block coordinate descent method for classification problems. This method separates the weights of the linear layer from the weights of the preceding nonlinear layers. NGD uses this decomposition to exploit the convexity of the cross-entropy loss with respect to the linear layer variables. It utilizes a Newton method to approximate the global minimum for a given batch of data before performing a step of gradient descent for the remaining variables. As such, it is a hybrid first/second order optimizer which extracts significant performance from a second-order substep that only scales with the number of weights in the linear layer, making it an appealing and feasible application of second-order methods for training very deep neural networks. Applying this optimizer to dense and convolutional networks, we have demonstrated acceleration with respect to the number of epochs in the validation loss for the peaks, MNIST, Fashion MNIST, and CIFAR-10 benchmarks, with improvements in accuracy for peaks benchmark and CIFAR-10 benchmark using a convolutional network.  

Examining the basis functions encoded in the hidden layer of the network in the peaks benchmarks revealed significant qualitative difference between NGD and stochastic gradient descent in the exploration of parameter space corresponding to the hidden layer variables. This, and the role of the tolerance in the Newton step as an implicit regularizer, merit further study. 

The difference in the regularity of the learned basis and probability classes suggests that one may obtain a qualitatively different model by varying only the optimization scheme used. We hypothesize that this more regular basis may have desirable robustness properties which may effect resulting model sensitivity. This could have applications toward training networks to be more robust against adversarial attacks.

\begin{ack}

Sandia National Laboratories is a multimission laboratory managed and operated by National Technology and Engineering Solutions of Sandia, LLC, a wholly owned subsidiary of Honeywell International, Inc., for the U.S. Department of Energy’s National Nuclear Security Administration under contract {DE-NA0003525}.  This paper describes objective technical results and analysis.  Any subjective views or opinions that might be expressed in the paper do not necessarily represent the views of the U.S. Department of Energy or the United States Government. SAND Number: {SAND2020-6022 J}. 

The work of R. Patel, N. Trask, and M. Gulian are supported by the U.S. Department of Energy, Office of Advanced Scientific Computing Research under the Collaboratory on Mathematics and Physics-Informed Learning Machines for Multiscale and Multiphysics Problems (PhILMs) project.  E. C. Cyr is supported by the Department of Energy early career program. M. Gulian is supported by the John von Neumann fellowship at Sandia National Laboratories.

\end{ack}

\bibliographystyle{abbrvnat} 
\bibliography{SNL_neurIPS}

\end{document}